\documentclass[11pt]{article}

\usepackage{chicagor}
\usepackage{times}
\usepackage{graphics}
\usepackage{amssymb}
\usepackage{amsmath}
\usepackage{proof}
\usepackage{theorems}

\let\Cite=\cite
\let\Citeyear=\citeyear

\allowdisplaybreaks[2]

\newenvironment{bprog}{\begin{array}[b]{@{}l@{}}}{\end{array}}

\newtheorem{theorem}{Theorem}[section]
\newtheorem{lemma}[theorem]{Lemma}

\newtheorem{EXAMPLE}[theorem]{Example}
\newtheorem{REMARK}[theorem]{Remark}

\newenvironment{example}{\begin{EXAMPLE} \rm}%
                            { \wbox\end{EXAMPLE}}
\newenvironment{example*}{\begin{EXAMPLE} \rm}%
                            {\end{EXAMPLE}}
                            { \wbox\end{REMARK}}
\newenvironment{remark*}{\begin{REMARK} \rm}%
                            {\end{REMARK}}

\def\squareforqed{\hbox{\rlap{$\sqcap$}$\sqcup$}}
\def\wbox{\ifmmode\squareforqed\else{\unskip\nobreak\hfil
\penalty50\hskip1em\null\nobreak\hfil\squareforqed
\parfillskip=0pt\finalhyphendemerits=0\endgraf}\fi}
\newenvironment{proof}{\noindent{\it Proof.}}
                      {\wbox\vspace{0.1in}}

\newcommand{\sat}{\models}

\renewcommand{\phi}{\varphi}

\renewcommand{\>}{\rangle}
\renewcommand{\emptyset}{\varnothing}
\newcommand{\teq}{\triangleq}

\newcommand{\hbra}{
\hbox to .995 \textwidth{\vrule width0.3mm height 1.8mm depth-0.3mm
                    \leaders\hrule height1.8mm depth-1.5mm\hfill
                    \vrule width0.3mm height 1.8mm depth-0.3mm}}
\newcommand{\hket}{
\hbox to .995 \textwidth{\vrule width0.3mm height1.5mm
                    \leaders\hrule height0.3mm\hfill
                    \vrule width0.3mm height1.5mm}}

\newcommand{\ratio}{.4}
\newenvironment{display}[1]{\begin{tabbing}
  \hspace{1.5em} \= \hspace{\ratio\linewidth-1.5em} \= \hspace{1.5em}
\= \kill
  {\bfseries#1}\\[-.8ex]
  \hbra\\[-.8ex]
  }{\\[-.8ex]\hket
  \end{tabbing}}
\newcommand{\entry}[2]{\>$#1$\>\>#2}
\newcommand{\clause}[2]{$#1$\>\>#2}
\newcommand{\category}[2]{\clause{#1::=}{#2}}
\newcommand{\categoryB}[2]{\clause{#1}{#2}}

\makeatletter
  \newcommand{\addToLabel}[1]{%
    \protected@edef\@currentlabel{\@currentlabel#1}%
  }
\makeatother

\newcounter{rule}

\newcommand{\staterule}[4][]{%
  \refstepcounter{rule}%
  \addToLabel{#2}%
  $\begin{array}[b]{@{}l}%
   \mbox{#2#1}\\%
   \begin{array}{c}
   #3\\
   \hline
   \raisebox{0ex}[2.5ex]{\strut}#4%
   \end{array}
  \end{array}$}

\newcommand{\GAP}{2ex}

\newcommand{\typerule}[4][]{%
  \refstepcounter{rule}%
  \addToLabel{#2}%
  $\begin{array}[b]{@{}l}%
   \mbox{#2#1}\\%
   \begin{array}{c}
   #3\\
   \hline
   \raisebox{0ex}[2.5ex]{\strut}#4%
   \end{array}
  \end{array}$}

\newcommand{\SUB}[1]{\{#1\}}
\newcommand{\GETS}{\mo{\leftarrow}}

\def\condarray#1#2#3{
    \left\{\begin{array}{ll}
        #2 & \mbox{if $#1$} \\
        #3 & \mbox{otherwise}
    \end{array}\right.}

\newcommand{\intension}[1]{[\![#1]\!]}

\newcommand{\typ}[1]{\mathsf{#1}}
\newcommand{\cS}{\mathcal{S}}
\newcommand{\cC}{\mathcal{C}}
\newcommand{\cO}{\mathcal{O}}

\newcommand{\con}[1]{\mathbf{#1}}

\newcommand{\fw}{/}
\newcommand{\bk}{\backslash}

\newcommand{\mo}{\mathord}

\newcommand{\true}{t\!t}
\newcommand{\false}{f\!\!f}

\newcommand{\Int}[1]{\mathbf{#1}}

\newcommand{\Class}[1]{\mathit{#1}}

\renewcommand{\phi}{\varphi}

\newcommand{\english}[1]{\textit{#1}}
\newcommand{\lexicon}[3]{\english{#1} \ensuremath{\mapsto {#2}:~{#3}}}

\newcommand{\trans}{\longrightarrow}

\newcommand{\Rule}[2]{\infer{#2}{#1}}

\title{A Framework for Creating Natural Language User Interfaces for
Action-Based Applications\thanks{A preliminary version of this paper
appeared in the \emph{Proceedings of the Third International AMAST Workshop on Algebraic
Methods in Language Processing}, TWLT Report 21, pp. 83-98, 2003.}}
\author{Stephen Chong\\
Cornell University\\
Ithaca, NY 14853 USA\\
schong@cs.cornell.edu
\and
Riccardo Pucella\\
Cornell University\\
Ithaca, NY 14853 USA\\
riccardo@cs.cornell.edu}
\date{}

\begin{document}
\maketitle

\begin{abstract}
In this paper we present a framework for creating natural language
interfaces to action-based applications.  
Our framework uses a number of reusable application-independent
components, in order to reduce the effort of creating a natural
language interface for a given application. Using a type-logical
grammar, we first translate natural language sentences into
expressions in an extended higher-order logic. These expressions can
be seen as executable specifications corresponding to the original
sentences. The executable specifications are then interpreted by
invoking appropriate procedures provided by the application for which
a natural language interface is being created. 
\end{abstract}

\section{Introduction}

The separation of the user interface from the application is regarded
as a sound design principle. A clean separation of these
components allows different user interfaces 
such as GUI, command-line and voice-recognition
interfaces. 
To support this feature, an application would supply an
\emph{application interface}. 
Roughly speaking, an application interface is a set of ``hooks'' that
an application provides so that user interfaces can access the
application's functionality. 
A user interface issues commands and queries to the application
through the application interface; the application executes these
commands and queries, and returns the results back to the user
interface.  We are interested in applications whose interface can be
described in terms of actions that modify the application's state, and
predicates that query the current state of the application. We refer
to such applications as \emph{action-based applications}.

\begin{figure}[t]
\begin{center}
\includegraphics{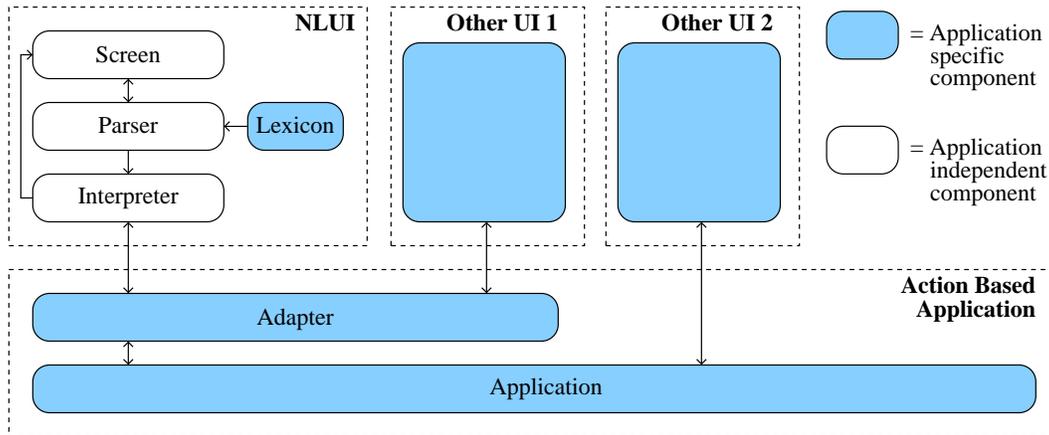}
\end{center}
\caption{Architecture}
\label{fig:architecture}
\end{figure}

In this paper, 
we propose a framework for creating natural language user interfaces
to action-based applications.   
These user interfaces will accept commands from the user in the form
of natural language sentences. 
We do not address how the user inputs these sentences (by
typing, by speaking into a voice recognizer, etc), but rather focus on
what to do with those sentences. Intuitively, we translate natural
language sentences into appropriate calls to procedures available
through the application interface. 

As an example, consider the application \textsc{ToyBlocks}. It
consists of a graphical representation of two blocks on a table, that
can be moved, and put one on top of the other. We would like to be
able to take a sentence such as \emph{move block one on block two},
and have it translated into suitable calls to the \textsc{ToyBlocks}
interface that would move block 1 on top of block 2. (This requires
that the interface of \textsc{ToyBlocks} supplies a procedure for moving
blocks.) While this example is simple, it already exposes most of the
issues with which our framework must deal.

Our framework architecture is sketched in
Figure~\ref{fig:architecture}.  The diagram shows an application with
several different user interfaces. The box labeled ``NLUI'' represents
the natural language user interface that our framework is designed to
implement. Our framework is appropriate for applications that provide
a suitable application interface, which is described in
Section~\ref{sec:action-based-app}. We expect that most existent
applications will not provide an interface conforming to our
requirements. Thus, an \emph{adapter} might be required, as shown in
the figure.  Other user interfaces can also build on this application
interface. The user interface labeled ``Other UI 1'' (for instance, a
command-line interface) does just that. The application may have
some user interfaces that interact with the application through other
means, such as the user interface ``Other UI 2'' (for instance, the
native graphical interface of the application).

The translation from natural language sentences to application
interface calls is achieved in two steps. The first step is to use a
categorial grammar~\Cite{r:carpenter97} to derive an intermediate
representation of the semantics of the input sentence. 
An interesting feature of categorial
grammars is that the semantics of the sentence is compositionally
derived from the meaning of the words in the lexicon. 
The derived meaning is a formula of higher-order logic~\Cite{r:andrews86}. 
The key observation is that such a formula
can be seen as an \emph{executable specification}. More precisely, it
corresponds to an expression of a simply-typed
$\lambda$-calculus~\Cite{r:barendregt81}. 
The second step
of our translation is to execute this $\lambda$-calculus expression
via calls to
procedures supplied by 
the application interface. 

We implement the above scheme as follows. 
A parser accepts a natural language
sentence from the user, and attempts to parse it using the
categorial grammar rules and the vocabulary from the application-specific lexicon. 
The parser fails if it is
not able to provide a unique unambiguous parsing of the
sentence. Successful parsing results in a formula in our higher-order
logic,
which corresponds to an expression in an action calculus---a
$\lambda$-calculus equipped with a notion of action.  
This expression is passed to the 
action calculus
interpreter, which ``executes'' the expression by making appropriate
calls to the application via the application interface. The
interpreter may report back to the screen the results of executing the
actions. 

The main 
advantage of our approach is its modularity.  This architecture
contains only a few application-specific components, and has a number
of reusable components. 
More precisely, the categorial grammar parser and the action calculus
interpreter are generic and reusable across different applications. 
The lexicon,
on the other hand,
 provides an application-specific vocabulary, and describes the
 semantics of the vocabulary in terms of a specific application
 interface.

In Section~\ref{sec:action-based-app} we describe 
our requirements for action-based applications. We define the notion
of an application interface, and provide a semantics for such an
interface in terms of a model of the application.
In Section~\ref{sec:nli} we present an
action calculus
that can be used to capture the meaning of imperative natural language sentences.
The semantics of this action calculus are given in terms of an
action-based application interface and application model; these
semantics permit us to evaluate expressions of the action calculus by
making calls to the application interface.
Section~\ref{sec:cat-gram} provides a brief introduction to
categorial grammars. Section~\ref{sec:all} shows how these components
(action-based applications, action calculus, and categorial
grammar) are used in our framework. We discuss some extensions to the
framework in Section~\ref{sec:extension}, and conclude in
Section~\ref{sec:conclusion}.

\section{Action-Based Applications}
\label{sec:action-based-app}

Our framework applies to applications that provide a
suitable Application Programmer Interface (API).  Roughly speaking,
such an interface provides procedures that are callable from external
processes to ``drive'' the application. In this section, we describe
in detail the kind of interface needed by our approach. We also
introduce a model of applications that will let us reason about the
suitability of the whole framework.

\subsection{Application Interface}
\label{sec:api}

Our framework requires action-based applications to have an
application interface that 
specifies which externally callable procedures exist in the
application.  
This interface is meant to
specify 
procedures that can be called from programs written in fairly
arbitrary programming languages. To achieve this, we assume only that
the calling language can distinguish between objects (the term
`object' is used is a nontechnical sense, to denote arbitrary data
values), and Boolean values $\true$ (true) and $\false$ (false). 

An application interface specifies the existence of a number of
different kind of procedures. 
\begin{enumerate}
\item {\bf Constants:} There is a set of constants representing
objects of interest. For \textsc{ToyBlocks}, the constants are
$\Int{b1}$, $\Int{b2}$, and $\Int{table}$. 

\item {\bf Predicates:} There is a set of predicates defined over the
states of the application. A predicate can be used to check whether
objects satisfy certain properties, dependent on the state of the
application. 
Predicates return truth values. 
For \textsc{ToyBlocks}, we consider the single predicate
$\Int{is\_on}(\mathit{bl},\mathit{pos})$, that checks whether a particular
block $\mathit{bl}$ is in a particular position $\mathit{pos}$ (on another
block or on the table). 
Each predicate $p$ has an associated
arity, indicating how many arguments it needs. 

\item {\bf Actions:} Finally, there is a set of actions defined by the
application. Actions essentially effect a state change. Actions can be
given arguments, for example, to effect a change to a particular
object. For \textsc{ToyBlocks}, we consider a single action,
$\Int{move}(\mathit{bl},\mathit{pos})$, which moves block $\mathit{bl}$ to
position $\mathit{pos}$ (on another block or on the table).  As with
predicates, each action has an associated arity, which may be $0$,
indicating that the action is parameterless.

\end{enumerate}

We emphasize that the application interface simply gives the names
of the procedures that are callable by external processes. It does not
actually define an implementation for these procedures. 

In order to prevent predicates and 
actions from being given inappropriate arguments, we need some
information about the actual kind of objects 
associated with constants, 
and that the predicates and actions take as arguments. 
We make
the assumption that every object in the application belongs to at least one of
many \emph{classes} of objects. Let $\cC$ be such a set of classes.
Although this terminology evokes object-oriented programming, we
emphasize that an object-oriented approach is not necessary for such
interfaces; a number of languages and paradigms are suitable for
implementing application interfaces.

We associate ``class information'' to every 
name in 
the interface
via a map $\sigma$. More specifically, we associate with every
constant $c$ a set $\sigma(c)\subseteq\cC$ representing the classes of
objects that can be associated with $c$. We associate with each
predicate $p$ a set $\sigma(p)\subseteq \cC^n$ (where $n$ is the arity
of the predicate), indicating for which classes of objects the
predicate is defined. Similarly, we associate with each action $a$ a
set $\sigma(a)\subseteq\cC^n$ (again, where $n$ is the arity of the
action, which in this case can be $0$). As we will make clear shortly,
we only require that the application return meaningful values for
objects of the right classes.

Formally, an application interface is a tuple $I=(C,P,A,\cC,\sigma)$,
where $C$ is a set of constant names, $P$ is a set of predicate names,
$A$ is a set of action names, $\cC$ is the set of classes of the
application, 
and $\sigma$ is the map associating every element of the interface
with its corresponding class information. 
The procedures in the interface provide a
means for an external process to access the functionality of the
application, by presenting to the language a generally accessible
version of the constants, predicates, and actions. 
Of course, in our case, we are not
interested in having arbitrary processes invoking procedures in the
interface, but specifically an interpreter that interprets commands
written in a natural language.

In a precise sense, the map $\sigma$ 
describes
typing information for the elements of the interface. However, because
we do not want to impose a particular type system on the application
(for instance, we do not 
want to assume that the application is object-oriented), we
instead assume a form of dynamic typing. More precisely, we
assume that there is a way to check if an object belongs to a given
class. This can either be performed through special {\em guard predicates} in the
application interface (for instance, a procedure $\Int{is\_block}$
that returns true if 
the supplied
object is actually a block), or a mechanism similar to Java's
\textit{instanceOf} operator.  

\begin{example}\label{x:ifc}
As an example, consider the following interface $I_T$ for
\textsc{ToyBlocks}. Let $I_T=(C,P,A,\cC,\sigma)$, where, as we
discussed earlier, 
\begin{itemize}
\item[] $\begin{array}{l}
C = \{\Int{b1},\Int{b2},\Int{table}\}\\
P = \{\Int{is\_on}\}\\
A = \{\Int{move}\}.
\end{array}$
\end{itemize}
We consider only two classes of objects, $\Class{block}$, representing
the blocks that can be moved, and $\Class{position}$, representing
locations where blocks can be located. Therefore,
$\cC=\{\Class{block},\Class{position}\}$. 

To define $\sigma$, consider the way in which the interface could be
used. The constant $\Int{b1}$ represents an object that is both a block that can be
moved, and a position to which the other block can be moved to (since we
can stack blocks on top of each other). The constant $\Int{b2}$  is similar.
The constant $\Int{table}$ represents an object that is a position only.
Therefore, we have:
\begin{itemize}
\item[] $\begin{array}{l}
  \sigma(\Int{b1}) = \{\Class{block},\Class{position}\}\\
  \sigma(\Int{b2}) = \{\Class{block},\Class{position}\}\\
  \sigma(\Int{table}) = \{\Class{position}\}.
  \end{array}$
\end{itemize}
Correspondingly, we can derive the class information for
$\Int{is\_on}$ and $\Int{move}$: 
\begin{itemize}
\item[] $\begin{array}{l}
  \sigma(\Int{is\_on}) = \{(\Class{block},\Class{position})\}\\
  \sigma(\Int{move}) = \{(\Class{block},\Class{position})\}.
  \end{array}$
\end{itemize}
\end{example}

\subsection{Application Model}
\label{sec:model}

In order to reason formally about the interface, we 
provide a semantics to the procedures in the interface. This 
is
done by supplying a model of the underlying application.  We make a
number of simplifying assumptions about the application model, and
discuss relaxing some of these assumptions in
Section~\ref{sec:extension}.

Applications are modeled using four components:

\begin{enumerate}

\item {\bf Interface:} The interface, as we saw in the previous
section, specifies the procedures that can be used to query and affect the
application. The interface also defines the set $\cC$ of classes of
objects in the application. 

\item {\bf States:} 
A state is, roughly speaking, everything that is relevant to
understand how the application behaves.  At any given point in time,
the application is in some state. We assume that an application's
state changes only through explicit actions.

\item {\bf Objects:} This defines the set of objects that can be
manipulated, or queried, in the application. As we already mentioned,
we use the term `object' in the generic sense, without implying that
the application is implemented through an object-oriented
language. Every object is associated with at least one class.

\item {\bf Interpretation:} An interpretation associates with every
element of the interface a ``meaning'' in the application model. As we
shall see, it associates with every constant an object of the model,
with every predicate a predicate on the model, and with every action a
state-transformation on the model. 

\end{enumerate}

Formally, an application is a tuple $M=(I,\cS,\cO,\pi)$, where $I$ in
an interface (that defines the constants, predicates, and actions of
the application, as well as the classes of the objects), $\cS$ is the
set of states of the application, $\cO$ is the set of objects, and
$\pi$ is the interpretation. 

We extend the map $\sigma$ defined in the interface to also provide
class information for the objects in $\cO$. Specifically, we define for
every object $o\in\cO$ a set $\sigma(o)\subseteq\cC$ of classes to
which that object belongs. 
An object can belong to more than one class. 

The map $\pi$ associates with every state and every element in the
interface (i.e., every constant, predicate and action)
the appropriate interpretation of that element at that
state. Specifically, for a state $s\in\cS$, 
we have $\pi(s)(c)\in\cO$. Therefore, constants can denote different
objects at different states of the applications. For predicates,
$\pi(s)(p)$ is a partial function from $\cO\times\ldots\times\cO$ to
truth values $\true$ or $\false$. This means that predicates 
are \emph{pure}, in that they do not modify the state of an
application; they are simply used to query the state. For actions, 
$\pi(s)(a)$ is a partial function from $\cO\times\ldots\times\cO$ to
$\cS$.  
The interpretation $\pi$ is subject to the following conditions. 
For a given predicate $p$, the interpretation $\pi(s)(p)$ must be
defined on objects of the appropriate class. Thus, the domain of the
partial function $\pi(s)(p)$ must at least consist of 
$\{(o_1, \ldots, o_n) ~|~ \sigma(o_1)\times\ldots\times\sigma(o_n)\cap\sigma(p)\not=\emptyset\}$. 
Similarly,
for a given action $a$, the domain of the partial function $\pi(s)(a)$
must at least consist of 
$\{(o_1, \ldots, o_n) ~|~ \sigma(o_1)\times\ldots\times\sigma(o_n) \cap \sigma(a) \not= \emptyset\}$. 
Furthermore, any class associated with a constant must also be
associated with the corresponding object.  In other words, for all
constants $c$, we must have $\sigma(c)\subseteq\sigma(\pi(s)(c))$
for all states $s$. 

\begin{example}\label{x:model}
We give a model $M_T$ for our sample \textsc{ToyBlocks} application,
to go with the interface $I_T$ defined in Example~\ref{x:ifc}. Let
$M_T=(I_T,\cS,\cO,\pi)$. We will consider only three states in the
application, $\cS=\{s_1,s_2,s_3\}$, which can be described variously: 
\begin{itemize}
\item[] \begin{tabular}{l}
in state $s_1$, blocks 1 and 2 are on the table\\
in state $s_2$, block 1 is on block 2, and block 2 is on the table\\
in state $s_3$, block 1 is on the table, and block 2 is on block 1.
	\end{tabular}
\end{itemize}
We consider only three objects in the model, $\cO=\{b_1,b_2,t\}$,
where $b_1$ is block 1, $b_2$ is block 2, and $t$ is the table. We
extend the map $\sigma$ in the obvious way: 
\begin{itemize}
\item[] $\begin{array}{l}
\sigma(b_1) = \{\Class{block},\Class{position}\}\\
\sigma(b_2) = \{\Class{block},\Class{position}\}\\
\sigma(t) = \{\Class{position}\}.
	 \end{array}$
\end{itemize}
The interpretation for constants is particularly simple, as the
interpretation is in fact independent of the state (in other words,
the constants refer to the same objects at all states):
\begin{itemize}
\item[] $\begin{array}{l}
\pi(s)(\Int{b1}) =  b_1\\
\pi(s)(\Int{b2}) = b_2\\
\pi(s)(\Int{table}) = t.
	 \end{array}$
\end{itemize}
The interpretation of the $\Int{is\_on}$ predicates is straightforward:
\begin{itemize}
\item[] $\begin{array}{l}
\pi(s_1)(\Int{is\_on})(x) =  \left\{
  \begin{array}{ll}
  \true & \mbox{if $x\in\{(b_1,t),(b_2,t)\}$}\\
  \false & \mbox{if $x\in\{(b_1,b_1),(b_1,b_2),(b_2,b_1),(b_2,b_2)\}$}
  \end{array}\right.\\
\pi(s_2)(\Int{is\_on})(x) = \left\{
  \begin{array}{ll}
  \true & \mbox{if $x\in\{(b_1,b_2),(b_2,t)\}$}\\
  \false & \mbox{if $x\in\{(b_1,t),(b_1,b_1),(b_2,b_1),(b_2,b_2)\}$}
  \end{array}\right.\\
\pi(s_3)(\Int{is\_on})(x) = \left\{
  \begin{array}{ll}
  \true & \mbox{if $x\in\{(b_1,t),(b_2,b_1)\}$}\\
  \false & \mbox{if $x\in\{(b_1,b_1),(b_1,b_2),(b_2,t),(b_2,b_2)\}$.}
  \end{array}\right.
	 \end{array}$
\end{itemize}
The interpretation of $\Int{move}$ is also straightforward:
\begin{itemize}
\item[] $\begin{array}{l}
\pi(s_1)(\Int{move})(x) = \left\{
  \begin{array}{ll}
  s_2 & \mbox{if $x=(b_1,b_2)$}\\
  s_3 & \mbox{if $x=(b_2,b_1)$}\\
  s_1 & \mbox{if $x\in\{(b_1,t),(b_1,b_1),(b_2,t),(b_2,b_2)\}$}
  \end{array}\right.\\
\pi(s_2)(\Int{move})(x) = \left\{
  \begin{array}{ll}
  s_1 & \mbox{if $x=(b_1,t)$}\\
  s_2 & \mbox{if $x\in\{(b_1,b_1),(b_1,b_2),(b_2,t),(b_2,b_1),(b_2,b_2)\}$}
  \end{array}\right.\\
\pi(s_3)(\Int{move})(x) = \left\{
  \begin{array}{ll}
  s_1 & \mbox{if $x=(b_2,t)$}\\
  s_3 & \mbox{if $x\in\{(b_1,t),(b_1,b_1),(b_1,b_2),(b_2,b_1),(b_2,b_2)\}$.}
  \end{array}\right.
	 \end{array}$
\end{itemize}
If a block is unmovable (that is, 
if there is another block on it), 
then the state does not change following a move operation.
\end{example}

\section{An Action Calculus}
\label{sec:nli}

Action-based application interfaces are designed to provide a means
for external processes to access the functionality of an
application. In this section we define a powerful and flexible
language that can be interpreted as calls to an application interface.
The language we use is a simply-typed $\lambda$-calculus extended with
a notion of action. It is effectively a computational
$\lambda$-calculus in the style of Moggi \Citeyear{r:moggi89}, although
we give a nonstandard presentation
in order to simplify expressing the language semantics in terms of an application interface.

The calculus is parameterized by a particular application interface
and application model. The application interface provides the
primitive constants, predicates, and actions, 
that can be used to build more complicated expressions, 
while the application model is used to define the semantics.

\subsection{Syntax}

Every expression in the language is given a type, intuitively
describing the kind of values that the expression produces. 
The types used in this language are given by the following grammar.
\begin{display}{Types:}
\category{\tau}{type}\\
\entry{\typ{Obj}}{object}\\
\entry{\typ{Bool}}{boolean}\\
\entry{\typ{Act}}{action}\\
\entry{\tau_1\rightarrow\tau_2}{function}
\end{display}
The types $\tau$ correspond closely to the types required by the 
action-based application interfaces we defined in the previous
section: the type $\typ{Bool}$ 
is the type of truth values, 
with constants $\con{true}$ and $\con{false}$ corresponding to the
Boolean values $\true$ and $\false$,
and the type $\typ{Obj}$ 
is the type of generic objects. 
The type $\typ{Act}$ is more subtle;
an expression of type $\typ{Act}$ represents an action 
that can be executed to change the state of the application. 
This is an example of computational type as defined by Moggi \Citeyear{r:moggi89}. 
As we shall see shortly, expressions of type $\typ{Act}$ can be
interpreted as calls to the action procedures of the application
interface.

The classes $\cC$ defined by the application interface have
no corresponding types in this language---instead, all objects have
the type $\typ{Obj}$. Incorporating these classes as types is 
an obvious 
possible extension (see Section~\ref{sec:extension}).

The syntax of the language 
is a straightforward extension of that of the $\lambda$-calculus. 

\begin{display}{Syntax of Expressions:}
\category{v}{value}\\
\entry{\con{true}~|~\con{false}}{boolean}\\
\entry{\lambda x\mo{:}\tau.e}{function}\\
\entry{\con{skip}}{null action}\\
\category{e}{expression}\\
\entry{x}{variable}\\
\entry{v}{value}\\
\entry{\mathit{id}_c()}{constant}\\
\entry{\mathit{id}_p(e_1,\ldots,e_n)}{predicate}\\
\entry{\mathit{id}_a(e_1,\ldots,e_n)}{action}\\
\entry{e_1~e_2}{application}\\
\entry{e_1?e_2\mo{:}e_3}{conditional}\\
\entry{e_1;e_2}{action sequencing}
\end{display}
The expressions $\mathit{id}_c()$, $\mathit{id}_p(e_1,\ldots,e_n)$ and
$\mathit{id}_a(e_1,\ldots,e_n)$ 
correspond to the procedures (respectively, constants, predicates,
and actions) available in the application
interface.
(Constants are written $\mathit{id}_c()$ as a
visual reminder that they are essentially functions: $\mathit{id}_c()$
may yield different values at different states, as the semantics will
make clear.)
So, for \textsc{ToyBlocks}, the constants are
$\con{b1}$, $\con{b2}$, and $\con{table}$; the only predicate
is $\con{is\_on}$; and the only action is $\con{move}$. 
The expression $e_1?e_2\mo{:}e_3$ is a
conditional expression, evaluating to $e_2$ if $e_1$ evaluates to
$\con{true}$, and $e_3$ if $e_1$ evaluates to $\con{false}$. 
The expression $e_1;e_2$ (when $e_1$ and $e_2$ are actions) evaluates
to an action corresponding to performing $e_1$ followed by $e_2$. The
constant $\con{skip}$ represents an action that has no effect.

\begin{example}\label{x:syntax}
Consider the interface for \textsc{ToyBlocks}. The expression
$\con{b1}()$ represents block 1, while $\con{table}()$ represents
the table. The expression $\con{move}(\con{b1}(),\con{table}())$
represents the action of moving block 1 on the table. Similarly, the
action
$\con{move}(\con{b1}(),\con{table}());\con{move}(\con{b2}(),\con{b1}())$
represents the composite action of moving block 1 on the table, and
then moving block 2 on top of block 1. 
\end{example}

\subsection{Operational Semantics}
\label{s:opsem}

The operational semantics is defined with respect to the application
model. More precisely, the semantics is given by a transition relation,
written $(s,e)\trans(s',e')$, where $s,s'$ are states of the
application, and $e,e'$ are expressions. Intuitively, this represents
the expression $e$ executing in state $s$, and making a one-step
transition to a (possibly different) state $s'$ and a new expression
$e'$. 

To accommodate the transition relation, we need to extend the syntax of
expressions to account for object values produced during the
evaluation. We also include a special value $\star$ that represents an
exception raised by the code. This exception 
is 
used to capture various errors that may occur during evaluation. 
\clearpage
\begin{display}{Additional Syntax of Expressions:}
\categoryB{v_o\in\cO}{object value}\\
\category{v}{value}\\
 \entry{...}\\
 \entry{v_o}{object}\\
 \entry{\star}{exception}
\end{display}

The transition relation is parameterized by the functions
$\delta_c$, $\delta_p$ and $\delta_p$, given below. These functions
provide a semantics to the constant, predicate, and action procedures
respectively, 
and are
derived from the interpretation $\pi$
in the application model.  
The intuition is
that evaluating these functions 
corresponds to 
making calls to the appropriate procedures on the given 
application interface, and returning the result. 
\begin{display}{Reduction Rules for Interface Elements:}
\clause{\delta_c(s,\mathit{id}_c) \teq \pi(s)(\mathit{id}_c)}\\
\clause{\delta_p(s,\mathit{id}_p, v_1, \ldots, v_n)  \teq 
  \condarray{\sigma(v_1)\times\ldots\times\sigma(v_n)\cap\sigma(\mathit{id}_p) \not= \emptyset}
            {\pi(s)(\mathit{id}_p)(v_1, \ldots, v_n)}
            {\star}}\\
\clause{\delta_a(s,\mathit{id}_a, v_1, \ldots, v_n) \teq
  \condarray{\sigma(v_1)\times\ldots\times\sigma(v_n)\cap\sigma(\mathit{id}_a)\not=\emptyset}
            {\pi(s)(\mathit{id}_a)(v_1, \ldots, v_n)}
            {\star}}
\end{display}
Note that 
determining whether or not a primitive throws an exception depends on
being able to establish the class of an object (via the map
$\sigma$). 
We can thus ensure that we never call an
action or predicate procedure on the application interface with
inappropriate objects, and so we guarantee a kind of dynamic
type-safety with respect to the application interface.

\begin{display}{Reduction Rules:}
\staterule
  {(Red App 1)}
  {(s,e_1)\trans(s,e_1')}
  {(s,e_1~e_2)\trans(s,e_1'~e_2)}
\staterule
  {(Red App 2)}
  {(s,e_1)\trans(s,\star)}
  {(s,e_1~e_2)\trans(s,\star)}
\staterule
  {(Red App 3)}
  {}
  {(s,(\lambda x\mo{:}\tau.e_1)~e_2)\trans(s,e_1\SUB{x\GETS e_2})}\\[\GAP]

\staterule
  {(Red OCon)}
  {}
  {(s,\mathit{id}_c())\trans(s,\delta_c(s,\mathit{id}_c))}
\staterule
  {(Red PCon 1)}
  {(s,e_i)\trans(s,e_i')\quad\mbox{for some $i\in[1..n]$}}
  {(s,\mathit{id}_p(\ldots,e_i,\ldots))\trans(s,\mathit{id}_p(\ldots,e_i',\ldots))}\\[\GAP]
\staterule
  {(Red PCon 2)}
  {(s,e_i)\trans(s,\star)\quad\mbox{for some $i\in[1..n]$}}
  {(s,\mathit{id}_p(e_1,\ldots,e_n))\trans(s,\star)}
\staterule
  {(Red PCon 3)}
  {}
  {(s,\mathit{id}_p(v_1,\ldots,v_n))\trans(s,v)}
  {$\delta_p(s,\mathit{id}_p,v_1,\ldots,v_n) = v$}\\[\GAP]

\staterule
  {(Red If 1)}
  {(s,e_1)\trans(s,e_1')}
  {(s,e_1?e_2\mo{:}e_3)\trans(s,e_1'?e_2\mo{:}e_3)}
\staterule
  {(Red If 2)}
  {(s,e_1)\trans(s,\star)}
  {(s,e_1?e_2\mo{:}e_3)\trans(s,\star)}
\staterule
  {(Red If 3)}
  {}
  {(s,v?e_{\con{true}}\mo{:}e_{\con{false}})\trans(s,e_v)} \\[\GAP]

\staterule
  {(Red Seq 1)}
  {(s,e_1)\trans(s',e_1')}
  {(s,e_1;e_2)\trans(s',e_1';e_2)}
\staterule
  {(Red Seq 2)}
  {}
  {(s,\star;e)\trans(s,\star)}
\staterule
  {(Red Seq 3)}
  {}
  {(s,\con{skip};e)\trans(s,e)} \\[\GAP]

\staterule
  {(Red ACon 1)}
  {(s,e_i)\trans(s,e_i')\quad \mbox{for some $i\in[1..n]$}}
  {(s,\mathit{id}_a(\ldots,e_i,\ldots))\trans(s,\mathit{id}_a(\ldots,e_i',\ldots))}
\staterule
  {(Red ACon 2)}
  {(s,e_i)\trans(s,\star)\quad\mbox{for some $i\in[1..n]$}}
  {(s,\mathit{id}_a(e_1,\ldots,e_n))\trans(s,\star)}\\[\GAP]
\staterule
  {(Red ACon 3)}
  {}
  {(s,\mathit{id}_a(v_1,\ldots,v_n))\trans(s',\con{skip})}
  {$\delta_a(s,\mathit{id}_a,v_1,\ldots,v_n) = s'$}
\\[\GAP]
\staterule
  {(Red ACon 4)}
  {}
  {(s,\mathit{id}_a(v_1,\ldots,v_n))\trans(s,\star)}
  {$\delta_a(s,\mathit{id}_a,v_1,\ldots,v_n) = \star$}
\end{display}
The operational semantics is a combination of call-by-name and
call-by-value semantics. The language as a whole is evaluated in a
call-by-name fashion. In particular, rule (Red App 3) indicates that
application is call-by-name. Actions, on the other hand, are evaluated
under what might be called call-by-value, as indicated by rule (Red
Seq 1). Roughly, the first term of a sequencing operation $e_1;e_2$ is
fully evaluated before $e_2$ is evaluated.  Intuitively, applications
are evaluated under call-by-name because premature evaluation of
actions could lead to action procedures in the application interface
being called inappropriately. For example, under call-by-value
semantics, the evaluation of the following expression 
\[(\lambda x \mo{:} \typ{Act}.
\con{false}? x \mo{:} \con{skip}) ~ \con{A}\]
would call the action procedure for $\con{A}$, assuming $\con{A}$ is
an action in the application interface.  This does not agree with the
intuitive interpretation of actions. More importantly, the mapping
from natural language sentences to expressions in our calculus
naturally yields a call-by-name interpretation.

\subsection{Type System}

\newcommand{\Jpure}[1]{\vdash #1~\mathrm{pure}}
\newcommand{\nJpure}[1]{\nvdash #1~\mathrm{pure}}
\newcommand{\Jimp}[1]{\vdash #1~\mathrm{imp}}
\newcommand{\Jok}[1]{\vdash #1~\mathrm{ok}}

We use type judgments to ensure that expressions are assigned types
appropriately, and that the types themselves are
\emph{well-formed}. Roughly speaking, a type is well-formed if it
preserves the separation between pure computations (computations with
no side-effects) and imperative computations (computations that may
have side-effects). 
The type system enforces that pure computations do not change the
state of the application. This captures the intuition that declarative
sentences---corresponding to pure computations--- should not change
the state of the world. (This correspondence between declarative
sentences and pure computations is made clear in the next section.)
The rules for the
type well-formedness judgment $\Jok{\tau}$ are given in
the following table, along with the auxiliary judgment
$\Jpure{\tau}$, stating that a type $\tau$ is a pure type (evaluates
without side effects). 

\clearpage
\begin{display}{Judgments $\Jpure{\tau}$ and $\Jok{\tau}$:} 
\typerule
  {(Pure Obj)}
  {}
  {\Jpure{\typ{Obj}}}
\typerule
  {(Pure Bool)}
  {}
  {\Jpure{\typ{Bool}}}
\typerule
  {(Pure Fun)}
  {}
  {\Jpure{\tau_1\rightarrow\tau_2}}
\typerule
  {(OK Fun Pure)}
  {\Jpure{\tau_1}\quad\Jpure{\tau_2}}
  {\Jok{\tau_1\rightarrow\tau_2}}
\typerule
  {(OK Fun Act)}
  {}
  {\Jok{\tau\rightarrow\typ{Act}}}
\end{display}

The judgment $\Gamma\vdash e:\tau$ assigns a type $\tau$ to expression
$e$ in a well-formed environment $\Gamma$. An environment $\Gamma$
defines the types of all variables in scope. An environment is of the
form $x_1:\tau_1,\ldots,x_n:\tau_n$, and defines each variable $x_i$
to have type $\tau_i$. We require that variables do not repeat in a
well-formed environment. The typing rules for expressions 
are essentially standard, with the exception of the typing rule for
functions, which requires that function types $\tau \rightarrow \tau'$
be well-formed.

\begin{display}{Judgment $\Gamma\vdash e:\tau$:}
\typerule
  {(Typ Var)}
  {}
  {\Gamma,x:\tau\vdash x:\tau}
\typerule
  {(Typ Obj)}
  {}
  {\Gamma\vdash v_o:\typ{Obj}}
\typerule
  {(Typ True)}
  {}
  {\Gamma\vdash \con{true}:\typ{Bool}}
\typerule
  {(Typ False)}
  {}
  {\Gamma\vdash \con{false}:\typ{Bool}}
\typerule
  {(Typ Exc)}
  {}
  {\Gamma\vdash \star:\tau}\\[\GAP]

\typerule
  {(Typ App)}
  {\Gamma\vdash e_1:\tau\rightarrow\tau' \quad \Gamma\vdash e_2:\tau}
  {\Gamma\vdash e_1~e_2:\tau'}
\typerule
  {(Typ Fun)}
  {\Gamma,x:\tau\vdash e:\tau'  \quad \Jok{\tau\rightarrow\tau'}}
  {\Gamma\vdash \lambda x\mo{:}\tau.e : \tau\rightarrow\tau'}
($x\not\in\mathrm{Dom}(\Gamma)$)\\[\GAP]

\typerule
  {(Typ If)}
  {\Gamma\vdash e_1:\typ{Bool} \quad \Gamma\vdash e_2:\typ{\tau}
    \quad \Gamma\vdash e_3:\typ{\tau}}
  {\Gamma\vdash e_1?e_2\mo{:}e_3:\typ{\tau}}
\typerule
  {(Typ Skip)}
  {}
  {\Gamma\vdash \con{skip}:\typ{Act}}
\typerule
  {(Typ Seq)}
  {\Gamma\vdash e_1:\typ{Act} \quad \Gamma\vdash e_2:\typ{Act}}
  {\Gamma\vdash e_1;e_2:\typ{Act}}\\[\GAP]

\typerule
  {(Typ ACon)}
  {\Gamma\vdash e_i:\typ{Obj} \quad \forall i\in[1..n]}
  {\Gamma\vdash\mathit{id}_a(e_1,\ldots,e_n):\typ{Act}}
\typerule
  {(Typ OCon)}
  {}
  {\Gamma\vdash\mathit{id}_c():\typ{Obj}}
\typerule
  {(Typ PCon)}
  {\Gamma\vdash e_i:\typ{Obj}\quad \forall i\in[1..n]}
  {\Gamma\vdash\mathit{id}_p(e_1,\ldots,e_n):\typ{Bool}}

\end{display}
It is straightforward to show that our type system is sound,
that is, that type-correct expressions do 
not get stuck when evaluating.
We write $(s,e)\trans^*(s',e')$ to mean that there exists a sequence
$(s_1,e_2),\dots,(s_n,e_n)$ such that
$(s,e)\trans(s_1,e_1)\trans\dots\trans(s_n,e_n)\trans(s',e')$.

\begin{theorem}\label{t:soundness}
If 
$\vdash e:\tau$, and $s$ is 
a state, then there exists a state $s'$ and value $v$ such that 
$(s,e)\trans^*(s',v)$. Moreover, if $\Jpure{\tau}$, then $s'=s$. 
\end{theorem}
\begin{proof}
See Appendix~\ref{a:proofs}.
\end{proof}

Theorem~\ref{t:soundness} in fact states that the language is strongly
normalizing: the evaluation of every expression 
terminates.
This is a very desirable property for the language, since it will form
part of the user interface.

\begin{example}\label{x:sem}
Consider the following example, interpreted with respect to the
application model of Example~\ref{x:model}. In state $s_1$ (where both
block 1 and 2 are on the table), let us trace through the execution of
the expression $(\lambda x\mo{:}\typ{Obj}.\lambda
y\mo{:}\typ{Obj}.\con{move}(x,y))~(\con{b1}())~(\con{b2}())$. (We
omit the derivation indicating how each step is justified.)
\[\begin{array}{l}
(s_1,(\lambda x\mo{:}\typ{Obj}.\lambda y\mo{:}\typ{Obj}.\con{move}(x,y)) ~ (\con{b1} ()) ~ (\con{b2}()))  
    \trans \\
 \quad (s_1,(\lambda y\mo{:}\typ{Obj}.\con{move}(\con{b1} (),y))  ~ (\con{b2}()))
    \trans \\
 \quad (s_1,\con{move}(\con{b1} (),\con{b2}()))
    \trans \\
 \quad (s_1,\con{move}(b_1,\con{b2}()))
    \trans \\
 \quad (s_1,\con{move}(b_1,b_2))  \trans \\
 \quad(s_2,\con{skip}).
  \end{array}\]
In other words, evaluating the expression in state $s_1$ leads to
state $s_2$, where indeed block 1 is on top of block 2. 
\end{example}

\subsection{A Direct Interpreter}

The main reason for introducing the action calculus of this section is
to provide a language in which to write expressions invoking
procedures available in the application interface. However, the
operational semantics given above rely on explicitly passing around
the state of the application. This state is taken from the application
model. In the model, the state is an explicit datum that enters the
interpretation of constants, predicates and actions. Of course, in the
actual application, the state is implicitly maintained by the
application itself. Invoking an action procedure on the application
interface modifies the current state of the application, putting the
application in a new state. This new state is not directly visible to
the user.

We can implement an interpreter based on the above operational
semantics but without carrying around the state explicitly.  To see
this, observe that the state is only relevant for the evaluation of
the primitives (constants, predicates, and actions). More importantly,
it is always the current state of the application that is relevant,
and only actions are allowed to change the state. We can therefore
implement an interpreter by simply directly invoking the procedures in
the application interface when the semantics tells us to reduce via
$\delta_c$, $\delta_p$, or $\delta_a$. Furthermore, we need to
be able to raise an exception $\star$ if the objects passed to the
interface are not of the right class. This requires querying for the
class of an object. As we indicated in Section~\ref{sec:api}, we
simply assume that this can be done, either through language
facilities (an \textit{instanceOf} operator), or through explicit
procedures in the interface that check whether an object is of a given
class.

In summary, given an application with a suitable application
interface, we can write an interpreter for our action calculus that
will interpret expressions by invoking procedures available through
the application interface when appropriate. The interpreter does not
require an application model. The model is useful to establish
properties of the interpreter, and if one wants to reason about the
execution of expressions via the above operational semantics.

\section{Categorial Grammars}
\label{sec:cat-gram}

In the last section, we introduced an action calculus that lets
us write expressions that can be understood via calls to the
application interface. The aim of this section is to use this action
calculus as the target of a translation from natural language
sentences. In other words, we describe a way to take a natural
language sentence and produce a corresponding expression in our action
calculus that captures the meaning of the sentence. Our main tool is
categorial grammars. 

Categorial grammars provide a mechanism to assign semantics to
sentences in natural language in a compositional manner. As we shall
see, we can obtain a compositional translation from natural language
sentences into the action calculus presented in the
previous section, and thus provide a 
simple natural language user interface for a given application. This section provides a brief
exposition of categorial grammars, based on Carpenter's
\Citeyear{r:carpenter97} presentation. 
We should note that the use of categorial grammars is not a
requirement in our framework. Indeed, any approach to provide
semantics to natural language sentences in higher-order logic, which
can be viewed as a simply-typed $\lambda$-calculus \Cite{r:andrews86},
can be adapted to our use. 
For instance, Moortgat's \Citeyear{r:moortgat97} multimodal categorial
grammars, which can handle a wider range of syntactic constructs, can
also be used for our purposes. To simplify the exposition, we use the
simpler categorial grammars in this paper. 

Categorial grammars were originally developed by Ajdukiewicz
\Citeyear{r:ajdukiewicz35} and Bar-Hillel \Citeyear{r:barhillel53}, and
later generalized by Lambek \Citeyear{r:lambek58}.  The idea behind
categorial grammars is simple. We start with a set of
\emph{categories}, each category representing a grammatical
function. For instance, we can start with the simple categories
\emph{np} representing noun phrases, \emph{pp} representing
prepositional phrases, \emph{s} representing declarative sentences and
\emph{a} representing imperative sentences. Given categories $A$ and
$B$, we can form the \emph{functor} categories $A \fw B$ and $B \bk
A$. The category $A \fw B$ represents the category of syntactic units
that take a syntactic unit of category $B$ to their right to form a
syntactic unit of category $A$. Similarly, the category $B \bk A$
represents the category of syntactic units that take a syntactic unit
of category $B$ to their left to form a syntactic unit of category
$A$.

\newcommand{\np}{\mathit{np}}
\newcommand{\pp}{\mathit{pp}}

Consider some examples. 
If $\np$ is the category of noun phrases and $s$ is the category of
declarative sentences, then 
the category $\np \bk s$ is the category of
intransitive verbs (e.g., \english{laughs}): they take a noun phrase
on their left to form a sentence (e.g., \english{Alice laughs} or
\english{the reviewer laughs}). Similarly, the category $(\np \bk s)
\fw \np$ represents the category of transitive verbs (e.g.,
\english{takes}): they take a noun phrase on their right and then a
noun phrase on their left to form a sentence (e.g., \english{Alice
takes the doughnut}).
We also consider the category $\pp$ of propositional phrases, as well
as the category $a$ of imperative sentences. 

The 
main
goal of categorial grammars 
is 
to provide a method of determining the well-formedness of natural
language. A \emph{lexicon} associates every word (or complex sequence
of words that constitute a single lexical entry) with one or more
categories. The approach described by Lambek \Citeyear{r:lambek58} is to
prescribe a calculus of categories so that if a sequence of words can
be assigned a category $A$ according to the rules, then the sequence
of words is deemed a well-formed syntactic unit of category
$A$. Hence, a sequence of words
is a well-formed noun phrase if it can be shown in the calculus that it
has category $np$. As an example of reduction, we see that if
$\sigma_1$ has category $A$ and $\sigma_2$ has category $A \bk B$,
then $\sigma_1~\sigma_2$ has category B. Schematically, $A, A \bk
B\Rightarrow B$. Moreover, this goes both ways, that is, if
$\sigma_1~\sigma_2$ has category $B$ and $\sigma_1$ can be shown to
have category $A$, then we can derive that $\sigma_2$ has category $A
\bk B$.

Van Benthem \Citeyear{r:benthem86} showed that this calculus could be
used to assign a semantics to terms by following the derivation of the
categories. 
Assume that every
basic category is assigned a type in our 
action calculus,
through a type assignment $T$. A type assignment $T$ can be extended
to functor categories by putting $T (A \fw B) = T (B \bk A) = T(B)
\rightarrow T(A)$. The lexicon is extended so that every word is now
associated with one or more pairs of a category $A$ and an expression $\alpha$
in our action calculus of the appropriate type,
that is, 
$\vdash \alpha : T(A)$.

We use the sequent notation $\alpha_1:A_1,\ldots,\alpha_n:A_n
\Rightarrow \alpha:A$ to mean that expressions
$\alpha_1,\ldots,\alpha_n$ of categories $A_1,\ldots,A_n$ can be
concatenated to form an expression $\alpha$ of category $A$. 
We call $\alpha:A$ the conclusion of the sequent. 
We 
use capital Greek letters ($\Sigma,\Delta$,...) to represent sequences
of expressions and categories. (We reserve $\Gamma$ for typing
contexts of the calculus in the last section.)  
We now give rules that allow us to derive new sequents from other
sequents.

\begin{display}{Categorial Grammar Sequent Rules:}
\typerule
  {(Seq Id)}
  {}
  {\alpha:A\Rightarrow \alpha:A} 
\typerule
  {(Seq Cut)}
  {\Delta\Rightarrow \beta:B \quad \Sigma_1,\beta:B,\Sigma_2\Rightarrow\alpha:A}
  {\Sigma_1,\Delta,\Sigma_2\Rightarrow \alpha:A}\\[\GAP]
\typerule
  {(Seq App Right)}
  {\Delta\Rightarrow \beta:B \quad \Sigma_1,\alpha(\beta):A,\Sigma_2\Rightarrow\gamma:C}
  {\Sigma_1,\alpha:A \fw B,\Delta,\Sigma_2\Rightarrow \gamma:C} 
\typerule
  {(Seq App Left)}
  {\Delta\Rightarrow \beta:B \quad \Sigma_1,\alpha(\beta):A,\Sigma_2\Rightarrow\gamma:C}
  {\Sigma_1,\Delta,\alpha:B \bk A,\Sigma_2\Rightarrow\gamma:C}\\[\GAP]
\typerule
  {(Seq Abs Right)}
  {\Sigma,x:A\Rightarrow \alpha:B}
  {\Sigma\Rightarrow \lambda x.\alpha:B \fw A}
\typerule
  {(Seq Abs Left)}
  {x:A,\Sigma\Rightarrow \alpha:B}
  {\Sigma\Rightarrow \lambda x.\alpha:A \bk B}
\end{display}

\begin{example}\label{x:lexicon}
Consider the following simple lexicon, suitable for the
\textsc{ToyBlocks} application. The following types are associated
with the basic grammatical units: 
\begin{itemize}
\item[] $\begin{array}{l}
T(np) = \typ{Obj}\\
T(pp) = \typ{Obj}\\
T(s) = \typ{Bool}\\
T(a) = \typ{Act}.\end{array}$
\end{itemize}
Here is a lexicon that captures a simple input language for
\textsc{ToyBlocks}: 
\begin{itemize}
\item[] \begin{tabular}{l}
\lexicon{block one}{\con{b1}()}{\np}\\
\lexicon{block two}{\con{b2}()}{\np}\\
\lexicon{the table}{\con{table}()}{\np} \\
\lexicon{on}{(\lambda x\mo{:}\typ{Obj}.x)}{\pp \fw \np}\\
\lexicon{is}{(\lambda x\mo{:} \typ{Obj}. \lambda y\mo{:}\typ{Obj}. \con{is\_on}(y,x) )}{(\np \bk s ) \fw \pp} \\
\lexicon{if}{(\lambda x\mo{:} \typ{Bool}. \lambda y\mo{:}\typ{Act}. x?y \mo{:} \con{skip} )}{(a \fw a) \fw s}\\
\lexicon{move}{(\lambda x\mo{:} \typ{Obj}. \lambda y\mo{:}\typ{Obj}. \con{move}(x,y) )}{(a \fw \pp ) \fw \np}.
\end{tabular}
\end{itemize}
This is a particularly simple lexicon, since every entry is
assigned a single term and category. 
It is also a very specialized lexicon, for the purpose of
illustration; our treatment of \english{is} is specific to the
\textsc{ToyBlocks} example. 

Using the above lexicon, the sentence \emph{move block one on block
two} can be associated with the string of expressions and categories
$\lambda x\mo{:}\typ{Obj}.\lambda
y\mo{:}\typ{Obj}.\con{move}(x,y):\mathit{(a\fw pp)\fw np}$,
$\con{b1}():\mathit{np}$, $\lambda x\mo{:}\typ{Obj}.x:\mathit{pp \fw
np}$, $\con{b2}():\mathit{np}$. 
The following derivation shows that this
concatenation yields an expression of category $\mathit{a}$. 
(For reasons of space, we have elided the type annotations in
$\lambda$-abstractions.) 
\[
\Rule{
  \Rule{}{
          \con{b2}()\mo{:}\np\Rightarrow\con{b2}()\mo{:}\np}
  &
  \Rule{\Rule{}{
                \con{b1}()\mo{:}\np\Rightarrow\con{b1}()\mo{:}\np}
     & 
     \qquad(\dagger)}
  {
    \begin{bprog}
        \lambda x.\lambda y.\con{move}(x,y)\mo{:}(a\fw \pp)\fw \np, 
        \con{b1}()\mo{:}\np, (\lambda x.x)~(\con{b2}())\mo{:}\pp
        \Rightarrow \\ \quad
        (\lambda x.\lambda y.\con{move}(x,y))~
        (\con{b1}())~((\lambda x.x)~(\con{b2}()))\mo{:}a
    \end{bprog}
  }
}
{
  \begin{bprog}
      \lambda x.\lambda y.\con{move}(x,y)\mo{:}(a\fw \pp)\fw \np, 
      \con{b1}()\mo{:}\np, \lambda x.x\mo{:}\pp \fw \np,
      \con{b2}()\mo{:}\np  
      \Rightarrow \\ \quad
      (\lambda x.\lambda y.\con{move}(x,y))~
      (\con{b1}())~((\lambda x.x)~(\con{b2}()))\mo{:}a
  \end{bprog}
}
\]
where the subderivation $(\dagger)$ is simply:
\[(\dagger): \begin{array}{c}
  \Rule{
     \Rule{}{\begin{bprog}
                 (\lambda x.x)~(\con{b2}())\mo{:}\pp\Rightarrow\\ \quad
                 (\lambda x.x)~(\con{b2}())\mo{:}\pp
	      \end{bprog}}
      & 
      \Rule{}
      {
        \begin{bprog}
          (\lambda x.\lambda y.\con{move}(x,y))~
          (\con{b1}())~((\lambda
           x.x)~(\con{b2}()))\mo{:}a\Rightarrow\\ \quad
          (\lambda x.\lambda y.\con{move}(x,y))~
          (\con{b1}())~((\lambda x.x)~(\con{b2}()))\mo{:}a
	\end{bprog}
      }
    }
    {
      \begin{bprog}
        (\lambda x.\lambda y.\con{move}(x,y))~(\con{b1}())\mo{:}a\fw \pp, 
        (\lambda x.x)~(\con{b2}())\mo{:}\pp
        \Rightarrow \\ \quad
        (\lambda x.\lambda y.\con{move}(x,y))~
        (\con{b1}())~((\lambda x.x)~(\con{b2}()))\mo{:}a.
      \end{bprog}
    }
\end{array}
\]
Hence,
the sentence is a well-formed imperative sentence. 
Moreover, the derivation shows that the meaning of the sentence
\emph{move block one on block two} is 
\[(\lambda x\mo{:}\typ{Obj}.\lambda
y\mo{:}\typ{Obj}.\con{move}(x,y))~(\con{b1}())~((\lambda
x\mo{:}\typ{Obj}.x)~(\con{b2}())).\] 
The execution of this expression,
similar to the one in Example~\ref{x:sem}, shows that the intuitive
meaning of the sentence is reflected by the execution of the
corresponding expression.
\end{example}

One might hope that the expressions derived through a categorial
grammar derivation are always valid expressions of our action
calculus. 
To ensure that this property holds, we must somewhat
restrict the kind of categories that can appear in a derivation. 
Let us say that a
derivation \emph{respects imperative structure} if for every category
of the form $A\bk B$ or $A\fw B$ that appears in the
derivation, we have $\Jok{T(A)\rightarrow T(B)}$.
Intuitively, a derivation respects imperative
structure if it cannot construct declarative sentences that depend on
imperative subsentences,
i.e., a declarative sentence cannot have any ``side effects.''
(For the lexicon in Example~\ref{x:lexicon}, a derivation respects
imperative structure if and only if every category
of the form $\mathit{a}\bk B$ or $B\fw\mathit{a}$ that appears in the
derivation is either $\mathit{a}\bk \mathit{a}$ or
$\mathit{a}\fw\mathit{a}$.)
We can show that all such derivations correspond to admissible typing
rules in the type system of the last section. (An admissible typing
rule is a rule that does not add derivations to the type system; anything
derivable using the rule can be derived without the rule.)
\begin{theorem}\label{t:safe-trans} 
If $\alpha_1:A_1,\dots,\alpha_n:A_n\Rightarrow \alpha:A$ has a
derivation that respects imperative structure, then the rule
\[ \Rule{\Gamma\vdash\alpha_1:T(A_1)\quad\dots\quad
\Gamma\vdash\alpha_n:T(A_n)}{\Gamma\vdash\alpha:T(A)} \]
is an admissible typing rule.   
\end{theorem}
\begin{proof}
See Appendix~\ref{a:proofs}.
\end{proof}

Note that if each expression $\alpha_i:A_i$ is 
taken from the lexicon, then we have $\vdash \alpha_i:T(A_i)$ by
assumption, and therefore Theorem~\ref{t:safe-trans} says that if
$\alpha_1:A_1,\dots,\alpha_k:A_k\Rightarrow\alpha:A$ has a derivation
that respects imperative structure, then $\vdash \alpha:T(A)$.

So, given a natural language imperative sentence from the user, we use
the lexicon to find the corresponding expressions and category pairs
$\alpha_1:A_1,\ldots,\alpha_n:A_n$, and then attempt to parse it, that
is, to find a derivation for the sequent
$\alpha_1:A_1,\ldots,\alpha_n:A_n \Rightarrow \alpha:\mathit{a}$ that
respects imperative structure.  If a
unique such derivation exists, then we have an unambiguous parsing of
the natural language imperative sentence, and moreover, the action calculus
expression $\alpha$ is the semantics of the imperative sentence.

\section{Putting It All Together}
\label{sec:all}

We now have the major components of our framework: a model for
action-based applications and interfaces to them; 
an action calculus 
which can be interpreted as calls to an application
interface; and the use of categorial grammars to create expressions
in our 
action calculus
from natural language sentences.

Let's see how our framework combines these components by considering
an end-to-end example for \textsc{ToyBlocks}. Suppose the user
inputs
the sentence \english{move block one on block two}
when blocks 1 and 2 are both on the table.
Our framework would process this sentence in the following steps. 

\begin{enumerate}
\item {\bf Parsing:} The \textsc{ToyBlocks} lexicon is used to parse
the sentence. Parsing succeeds only if there is a unique parsing of
the sentence
(via a derivation that respects imperative structure), otherwise the
parsing step fails,  
because the sentence was either ambiguous, contained unknown words or
phrases, or was ungrammatical.  In this example, there is only a
single parsing of the sentence (as shown in Example~\ref{x:lexicon}),
and the result is the following expression in our action calculus,
which has type $\typ{Act}$: 
\[
(\lambda x \mo{:} \typ{Obj}. \lambda y \mo{:}
\typ{Obj}. \con{move}(x,y) ) ~ (\con{b1}()) ~~ ((\lambda
x\mo{:}\typ{Obj}.x) ~ (\con{b2}())). 
\]

\item {\bf Evaluating:} The action calculus expression is evaluated
using 
a direct interpreter implementing 
the operational semantics of Section~\ref{sec:nli}. The
evaluation of the expression proceeds as follows. 
\[\begin{array}{l}
(s_1, (\lambda x \mo{:} \typ{Obj}. \lambda y \mo{:} \typ{Obj}. \con{move}(x,y) ) ~ (\con{b1}()) ~ ((\lambda x\mo{:}\typ{Obj}.x) ~ (\con{b2}())))
    \trans \\
 \quad (s_1, (\lambda y \mo{:} \typ{Obj}. \con{move}(\con{b1}(),y) ) ~ ((\lambda x\mo{:}\typ{Obj}.x) ~ (\con{b2}())))
    \trans \\
 \quad (s_1, \con{move}(\con{b1}(),(\lambda x\mo{:}\typ{Obj}.x) ~ (\con{b2}()) ))
    \trans \\
 \quad (s_1, \con{move}(b_1,(\lambda x\mo{:}\typ{Obj}.x) ~ (\con{b2}()) ))
    \trans \\
 \quad (s_1, \con{move}(b_1,(\con{b2}())))
    \trans \\
 \quad (s_1, \con{move}(b_1,b_2))
    \trans \\
 \quad(s_2,\con{skip}).
  \end{array}\]

In the process of this evaluation, several calls are generated to the
application interface. In particular, calls are made to determine the
identity of the object constants $\con{b1}$ and $\con{b2}$ as $b_1$
and $b_2$ respectively. Then, during the last transition, 
guard predicates such as 
$\con{is\_block}(b_1)$ and $\con{is\_position}(b_2)$ may be
called to ensure that $b_1$ and $b_2$ are of the appropriate classes
for being passed as arguments to $\con{move}$. Since the objects are
of the appropriate classes, the action $\con{move}(b_1,b_2)$ is
invoked via the application interface, and succeeds.

\item {\bf Reporting:} Following the evaluation of the expression,
some result must be reported back to the user. Our framework does not
detail what information is conveyed back to the user, but they must be
informed if an exception was raised during the evaluation of the expression.

In this example, no exception was raised, so what to report to the
user is at the discretion of the user interface. If the user interface had a
graphical depiction of the state of \textsc{ToyBlocks}, it may now
send queries to the application interface to determine the new state
of the world, and modify its graphical display appropriately.

\end{enumerate}

Let's consider what would happen if an exception ($\star$) was raised
during the evaluation phase. For example, consider processing the
sentence \english{move the table on block one}. The parsing phase
would succeed, as the sentence is grammatically correct. However, prior
to calling the action $\con{move}(t,b_1)$, the evaluation would
determine that the object $t$ does not belong to the class
$\Class{block}$ (by a guard predicate such as $\con{is\_block}(t)$
returning $\false$, or by some other mechanism). An exception would
thus be raised, and some information must be reported back to the user
during the reporting phase. Note that the framework has ensured that
the action $\con{move}(t,b_1)$ was not invoked on the application
interface.

\section{Extensions}
\label{sec:extension}

Several extensions to this framework are possible. There is a mismatch
of types in our framework. The application model permits a rich notion
of types: any object of the application may belong to one or more
classes. By contrast, our action calculus has a very simple notion
of types, assigning the type $\typ{Obj}$ to all objects, and not
statically distinguishing different classes of objects. The simplicity
of our action calculus is achieved at the cost of dynamic type
checking, which ensures that actions and predicates on the application
interface are invoked only with appropriate parameters. It would be
straightforward to extend the action calculus with a more refined
type system that includes a notion of subtyping, to model the
application classes.  Not only would this extension remove many, if
not all, of the dynamic type checks,
but it may also reduce the number of possible parses of natural
language sentences. The refined type system allows the semantics of
the lexicon entries to be finer-grained, and by considering these
semantics, some nonsensical parses of a sentence could be
ignored.  For example, in the sentence \english{pick up the book and
the doughnut and eat it} the referent of \english{it} could naively be
either \english{the book} or \english{the doughnut}; if the semantics
of \english{eat} require an object of the class $\typ{Food}$ and the
classes of \english{the book} and \english{the doughnut} are
considered, then the former possibility could be ruled out.

Another straightforward extension to the framework is to allow the
user to query the state by entering declarative sentences and treating
them as yes-no interrogative sentences. For example, \english{block
one is on the table?} This corresponds to accepting sequents of the
form $\alpha_1:A_1,\ldots,\alpha_n:A_n \Rightarrow \alpha:s$,
and executing the action calculus expression $\alpha$, which has
type $\typ{Bool}$. The categorial grammar could be extended to accept
other yes-no questions, such as \english{is block two on block one?}
A more interesting extension (which would require a correspondingly
more complex application model) is to allow hypothetical queries, such
as \english{if you move block one on block two, is block one on the
table?} This corresponds to querying \english{is block one on the
table?} in the state that would result if the action \english{move
block one on block two} were performed. This extension would bring our
higher-order logic 
(that is, our action calculus) 
closer to dynamic logic~\Cite{r:groenendijk91,r:harel00}.
It is not clear, however, how to derive a direct interpreter for such
an extended calculus. 

In Section~\ref{sec:model} we made some simplifying assumptions about
the application model. Chief among these assumptions was that an
application's state changes only as a result of explicit actions. This
assumption may be unrealistic if, for example, the application has
multiple concurrent users. We can however extend the framework to
relax this assumption. One way of relaxing it is to incorporate
transactions into the application model and application interface: the
application model would guarantee that within transactions, states
change only as a result of explicit actions, but if no transaction is
in progress then states may change arbitrarily. The evaluation of an
action calculus expression would then be wrapped in a transaction.

Another restriction we imposed was that predicates be pure. It is of
course technically possible to permit arbitrary state changes during
the evaluation of predicates. In fact, we can modify the operational
semantics to allow the evaluation of any expression to change
states. 
If done properly, 
the key property is still preserved: the evaluation of constants,
predicates or actions rely only on the current state, and all other
transitions do not rely on the state at all. Thus, the semantics
remains consistent with interpreting expressions using calls to the
application interface.
However, doing this would lose the intuitive meaning of natural
language sentences that do not contain actions; they should not change
the state of the world.  

\section{Conclusion}
\label{sec:conclusion}

We have presented a framework that simplifies the creation of simple
natural language user interfaces for action-based applications. The
key point of this framework is the use of a $\lambda$-calculus to
mediate access to the application. The $\lambda$-calculus we define is
used as a semantics for natural language sentences (via categorial
grammars), and expressions in this calculus are executed by issuing
calls to the application interface. The framework has a number of
application-independent components, reducing the amount of effort
required to create a simple natural language user interface for a
given application.

A number of applications have natural language interfaces
\Cite{r:winograd71,r:price00}, but they appear to be designed specifically
for the given application, rather than being a generic approach. A
number of methodologies and frameworks exist for natural language
interfaces for database queries (see Androutsopoulos et
al. \Citeyear{r:androutsopoulos95} for a survey), but we are not aware
of a framework for deriving natural language interfaces to general
applications in a principled manner.

While the framework presented here is useful for the rapid development
of simple natural language user interfaces, the emphasis is on simple.
Categorial grammars (and other techniques that use higher order logic
as the semantics of natural language)  
are limited in their ability to deal with the wide
and diverse phenomena that occur in English.  For example, additional
mechanisms outside of the categorial grammar, probably
application-specific, would be required to deal with discourse.
However, categorial grammars are easily extensible, by expanding the
lexicon, and many parts of the lexicon of a categorial grammar are
reusable in different applications, making it well-suited to a
framework for rapid development of natural language user interfaces. 

It may seem that a limitation of our framework is that it is only suitable for
applications for which we can provide an interface of the kind
described in Section~\ref{sec:action-based-app}---the action calculus 
of Section~\ref{sec:nli} is specifically designed to be interpreted as
calls to an action-based application. However, all the examples we
considered can be provided with such an interface. 
It is especially interesting to note that our definition of
action-based application interfaces is compatible with the notion of
interface for XML web services \Cite{r:barclay02}. This suggests that it
may be possible to derive a natural language interface to XML Web
Services using essentially the approach we 
advocate in this paper.

\section*{Acknowledgments}

Thanks to Eric Breck and Vicky Weissman for comments on earlier drafts of this paper. 
This work was partially supported by NSF under grant CTC-0208535, by
ONR under grants N00014-00-1-03-41 and N00014-01-10-511, and by the
DoD Multidisciplinary University Research Initiative (MURI) program
administered by the ONR under grant N00014-01-1-0795.

\appendix

\newcommand{\citem}[1]{\item[-]Case #1:}
\newcommand{\mi}[1]{\mathit{#1}}

\section{Proofs}\label{a:proofs}

The soundness and strong normalization (Theorem~\ref{t:soundness}) of
the calculus in Section~\ref{sec:nli} can be derived using
\emph{logical relations}, in a fairly standard way
\cite{r:winskel93}. In order to do this, we need some lemmas about
properties of the operational semantics.
\begin{lemma}\label{l:preservation}
If $\vdash e:\tau$ and $(s,e)\trans(s',e')$, then $\vdash e':\tau$.
\end{lemma}
\begin{proof}
This is a completely straightforward proof by induction on the
height of the typing derivation for $\vdash e:\tau$. 
\end{proof}

\begin{lemma}\label{l:third}
If $\vdash e:\tau$, $(s,e)\trans(s',e')$, and $\Jpure{\tau}$, then
$s'=s$. 
\end{lemma}
\begin{proof}
This result follows essentially by examination of the 
operational semantics rules, proceeding by induction on the structure
of $e$. 

\begin{itemize}
\citem{$e=x$} This case cannot arise, since $\vdash e:\tau$ cannot
hold with an empty context when $e$ is a variable. 

\citem{$e=v$} An inspection of the operational semantics rules shows
that this case cannot arise, since there is no $s'$ and $e'$ such that
$(s,e)\trans(s',e')$ if $e$ is a value. 

\citem{$e=\mathit{id}_c()$} By (Red OCon), we have
$(s,e)\trans(s,\delta_c(\mi{id}_c))$, and the state is unchanged,
irrespectively of $\tau$.

\citem{$e=\mathit{id}_p(e_1,\ldots,e_n)$} By examination of the
operational semantics rules, two cases arise. If every $e_i$ is a
value $v_i$, then
$(s,e)\trans(s,\delta_p(s,\mi{id}_p,v_1,\dots,v_n))$, with
$\tau=\typ{Bool}$ and $\Jpure{\tau}$, and the state is unchanged
during the transition, as required. Otherwise, there is at least one
$e_i$ that is not a value, and
$(s,e)\trans(s,\mi{id}_p(\dots,e',\dots))$ or
$(s,e)\trans(s,\star)$. Again, $\tau=\typ{Bool}$, so that
$\Jpure{\tau}$, and the state is unchanged during the transition, as
required.

\citem{$e=\mathit{id}_a(e_1,\ldots,e_n)$} If $\vdash e:\tau$, then
$\tau=\typ{Act}$, which is not a pure type, so there is nothing to
show for this case. 

\citem{$e=e_1~e_2$} By examination of the operational semantics rules,
two cases arise. If $e_1$ is a value, then it must be $\star$ or an
abstraction $\lambda x\mo{:}\tau'.e'$. In the former case,
$(s,e)\trans(s,\star)$. In the latter case,
$(s,e)\trans(s,e'\SUB{x\GETS e_2})$. In both cases, the state is
unchanged, irrespectively of the type $\tau$. If $e_1$ is not a value,
then from rule (Red App 1), we get $(s,e_1~e_2)\trans(s,e_1'~e_2)$ or
$(s,e_1~e_2)\trans(s,\star)$, and the state is unchanged,
irrespectively of the type $\tau$.

\citem{$e=e_1?e_2\mo{:}e_3$} By examination of the operational semantics
rules, we consider two cases. If $e_1$ is a value, then it must be
$\star$ or a Boolean value. In the former case,
$(s,e)\trans(s,\star)$. In the latter case, $(s,e)\trans(s,e_2)$ or
$(s,e)\trans(s,e_3)$, depending on whether $e_1$ is $\con{true}$ or
$\con{false}$. In both cases, the state is unchanged, irrespectively
of the type $\tau$. If $e_1$ is not a value, then from rule (Red If
1), we get $(s,e)\trans(s,e_1'?e_2\mo{:}e_3)$ or $(s,e)\trans(s,\star)$,
and the state is unchanged, irrespectively of the type $\tau$.

\citem{$e=e_1;e_2$} If $\vdash e:\tau$, then $\tau=\typ{Act}$, which
is not a pure type, so there is nothing to show for this case.
\end{itemize}
This completes the induction.
\end{proof}

We define, for each type $\tau$, a set $R_\tau$ of terms which
terminate in all states. Formally, for a base type $b$, either
$\typ{Obj}$, $\typ{Bool}$, or $\typ{Act}$, we take
\[ R_b=\{e\mid ~\vdash e: t, \forall s\exists
v\exists s'.(e,s)\trans^*(v,s')\}.\]
For a function type $\tau_1\rightarrow\tau_2$, we take
\[ R_{\tau_1\rightarrow\tau_2}=\{e\mid ~\vdash
e:\tau_1\rightarrow\tau_2, \forall s\exists
v\exists s'.(e,s)\trans^*(v,s'), \forall e'\in R_{\tau_1}.(e~e_1)\in
R_{\tau_2}\}.\]

\newcommand{\dom}{\mathrm{dom}}

We define a substitution operator $\gamma$ to be a partial map from
variables to expressions of the action calculus. Let $\dom(\gamma)$ be
the domain of definition of the partial map $\gamma$. Given a
context $\Gamma$, we write $\gamma\sat\Gamma$ if the domains of
$\gamma$ and $\Gamma$ are equal (a context $\Gamma$ can be understood
as a partial map from variables to types), and for all
$x\in\dom(\gamma)$, $\gamma(x)\in R_{\Gamma(x)}$, where $\Gamma(x)$ is
the type associated with $x$ in the context $\Gamma$. We extend
$\gamma$ to expressions, by taking $\hat{\gamma}(e)$
to be the expression resulting from replacing every variable $x$ in $e$ by
the expression $\gamma(x)$. Formally,
\begin{align*}
\hat{\gamma}(x) & = \begin{cases} 
  \gamma(x) & \text{if $x\in\dom(\gamma)$}\\
  x & \text{otherwise}
		    \end{cases}\\
\hat{\gamma}(\con{true}) & = \con{true}\\
\hat{\gamma}(\con{false}) & = \con{false}\\
\hat{\gamma}(\lambda x\mo{:}\tau.e) & = \lambda x\mo{:}\tau.\hat{\gamma}_x(e)\\
\hat{\gamma}(\con{skip}) & = \con{skip}\\
\hat{\gamma}(v_o) & = v_o\\
\hat{\gamma}(\star) & = \star\\
\hat{\gamma}(\mathit{id}_c()) & = \mathit{id}_c()\\
\hat{\gamma}(\mathit{id}_p(e_1,\dots,e_n) & = \mathit{id}_p(\hat{\gamma}(e_1),\dots,\hat{\gamma}(e_n))\\
\hat{\gamma}(\mathit{id}_a(e_1,\dots,e_n) & = \mathit{id}_a(\hat{\gamma}(e_1),\dots,\hat{\gamma}(e_n))\\
\hat{\gamma}(e_1~e_2) & = \hat{\gamma}(e_1)~\hat{\gamma}(e_2)\\
\hat{\gamma}(e_1?e_2\mo{:}e_3) & = \hat{\gamma}(e_1)?\hat{\gamma}(e_2)\mo{:}\hat{\gamma}(e_3)\\
\hat{\gamma}(e_1;e_2) & = \hat{\gamma}(e_1);\hat{\gamma}(e_2)
\end{align*}
where
$\hat{\gamma}_x$ is the same substitution map as $\gamma$, except that
it is undefined on variable $x$. 

\begin{lemma}\label{l:first}
If $\Gamma\vdash e:\tau$ and $\gamma\sat\Gamma$, then $\vdash
\hat{\gamma}(e):\tau$. 
\end{lemma}
\begin{proof}
This is a straightforward proof by induction on the height of the
typing derivation for $\Gamma\vdash e:\tau$. 
\end{proof}

\begin{lemma}\label{l:second}
If $\vdash e:\tau$, and for all $s$ there exists $s'$ and $e'\in R_\tau$ such
that $(s,e)\trans^*(s',e')$, then $e\in R_\tau$.
\end{lemma}
\begin{proof}
We prove this by induction on the structure of $\tau$. For a base type
$b$, the result is immediate by the definition of $R_b$. For
$\tau=\tau_1\rightarrow\tau_2$, assume $\vdash
e:\tau_1\rightarrow\tau_2$, and for all $s$, there exists the required
$s',e'$. For an arbitrary state $s$, let $s',e'$ be such that
$(s,e)\trans^*(s',e')$; since $e'\in R_{\tau_1\rightarrow\tau_2}$, we
have $(s',e')\trans^*(s'',v)$ for some $s''$ and value $v$. Thus,
$(s,e)\trans^*(s'',v)$. Finally, it remains to show that for all
$e''\in R_{\tau_1}$, we have $(e~e'')\in R_{\tau_2}$. By assumption, we
have $(e'~e'')\in R_{\tau_2}$. To apply the induction hypothesis and get
$(e~e')\in R_{\tau_2}$, we show that for all $s$, we have
$(s,e~e'')\trans^*(s,e'~e'')$. We proceed by induction on the length
of the derivation $(s,e)\trans^*(s',e')$. First, note that because
$\vdash e:\tau_1\rightarrow\tau_2$, which is a pure type, a
straightforward induction on the length of the derivation using
Lemma~\ref{l:third} shows that $s'=s$. If the length is $0$, then
$e=e'$, so the result is immediate. If the length is non-zero, then
$(s,e)\trans^*(s,e''')\trans(s,e')$. By the induction hypothesis,
$(s,e~e'')\trans^*(s,e'''~e'')$. Since $(s,e''')\trans(s,e')$, by rule
(Red App 1), $(s,e'''~e'')\rightarrow(s,e'~e'')$, so that
$(s,e~e'')\trans^*(s,e'~e'')$, as required. This establishes that
$(e~e'')\in R_{\tau_2}$.
\end{proof}

We can now prove the main result. 

\begin{namedtheorem}{Theorem}{\ref{t:soundness}}
If $\vdash e:\tau$, and $s$ is a state, then there exists a state $s'$
and value $v$ such that $(s,e)\trans^*(s',v)$. Moreover, if
$\Jpure{\tau}$, then $s'=s$. 
\end{namedtheorem}
\begin{proof}
Clearly, it is sufficient to show that $\vdash e:\tau$ implies $e\in
R_\tau$. To use induction, we prove the more general statement that
$\Gamma\vdash e:\tau$ and $\gamma\sat\Gamma$ implies
$\hat{\gamma}(e)\in R_\tau$. (The desired result follows by taking
$\gamma$ to be the empty substitution, and $\Gamma$ the empty
context.) We prove the general result by induction on the structure of
$e$.

\begin{itemize}

\citem{$e=x$} Assume $\Gamma\vdash x:\tau$, and
$\gamma\sat\Gamma$. We need to show that $\hat{\gamma}(x)\in R_\tau$.
Since $x$ is a variable, $\tau=\Gamma(x)$, and thus $x$ is in the
domain of $\Gamma$. Since $\gamma\sat\Gamma$, $\gamma(x)\in R_\tau$,
and $\hat{\gamma}(x)=\gamma(x)$ implies $\hat{\gamma}(x)\in R_\tau$,
as required. 

\citem{$e=\con{true},\con{false},\con{skip},\star,v_o$} Assume
$\Gamma\vdash e:b$, for the appropriate base type $b$, and
$\gamma\sat\Gamma$. We need to show that $\hat{\gamma}(e)=e\in
R_b$. Since $e$ is a value, than for all $s$, $(s,e)\trans^*(s,e)$, so 
$e\in R_b$, as required. 

\citem{$e=\lambda x\mo{:}\tau.e'$} This is the difficult case. Assume
that $\Gamma\vdash \lambda x\mo{:}\tau.e':\tau\rightarrow\tau'$, and
$\gamma\sat\Gamma$. We need to show that $\hat{\gamma}(\lambda
x\mo{:}\tau.e')=\lambda x\mo{:}\tau.\hat{\gamma}(e')\in
R_{\tau\rightarrow\tau'}$. This involves, following the definition of
$R_{\tau\rightarrow\tau'}$, establishing three facts. First, by
Lemma~\ref{l:first}, $\vdash\hat{\gamma}(\lambda
x\mo{:}\tau.e')$. Since $\hat{\gamma}(\lambda\mo{:}\tau.e')=\lambda
x\mo{:}.\hat{\gamma}(e')$ is a value, we immediately have that
$(s,\lambda x\mo{:}.\hat{\gamma}(e'))$ reduces to a value for all
states $s$. Finally, we need to show that for all $e''\in R_{\tau}$,
we have $((\lambda x\mo{:}.\hat{\gamma}(e'))~e'')\in R_{\tau'}$. Given 
$e''\in R_{\tau}$. By (Red App 3), for all $s$, $(s,(\lambda
x\mo{:}\tau.\hat{\gamma}(e'))~e'')\trans(s,e'\SUB{x \GETS e''})$. By
Lemma~\ref{l:second}, it suffices to show that $e'\SUB{x\GETS e''}\in
R_{\tau'}$ to show that $((\lambda x\mo{:}.\hat{\gamma}(e'))~e'')\in
R_{\tau'}$ (by taking $s'=s$). 

Define $\gamma'_x=\gamma_x[x\mapsto e'']=\gamma[x\mapsto e'']$ (since
$\gamma_x$ is just $\gamma$ expect undefined on variable
$x$). Clearly, $\hat{\gamma}(e')\SUB{x\GETS
e''}=\hat{\gamma}'_x(e')$. By assumption, we have $\Gamma\vdash
\lambda x\mo{:}\tau.e':\tau\rightarrow\tau'$, which means that
$\Gamma,x:\tau\vdash e':\tau'$. Now, $\gamma'_x\sat\Gamma,x:\tau$,
since $\gamma\sat\Gamma$ and $\gamma'_x(x)=e''\in R_\tau$, by
assumption. Applying the induction hypothesis yields that
$\hat{\gamma}'_x(e')\in R_{\tau'}$, as required. 

\citem{$e=\mathit{id}_c()$} Assume $\Gamma\vdash
\mathit{id}_c():\typ{Obj}$, and  $\gamma\sat\Gamma$. We need to show
that $\hat{\gamma}(\mathit{id}_c())=\mathit{id}_c()\in
R_{\typ{Obj}}$. For all $s$,
$(s,\mathit{id}_c())\trans^*(s,\delta_c(s,\mathit{id}_c))$ by (Red
OCon), so $\mathit{id}_c()\in R_{\typ{Obj}}$, as required.

\citem{$e=\mathit{id}_p(e_1,\ldots,e_n)$} Assume
$\Gamma\vdash\mathit{id}_p(e_1,\dots,e_n):\typ{Bool}$, and
$\gamma\sat\Gamma$. We need to show that
$\hat{\gamma}(\mathit{id}_p(e_1,\ldots,e_n))=\mathit{id}_p(\hat{\gamma}(e_1),\dots,\hat{\gamma}(e_n))\in
R_{\typ{Bool}}$. Since $\Gamma\sat
\mathit{id}_p(e_1,\ldots,e_n):\typ{Bool}$, we have $\Gamma\sat
e_i:\typ{Obj}$ for all $i$. Applying the induction hypothesis, we get
that $e_i\in R_{\typ{Obj}}$ for all $i$, and thus for all $s$, we can
construct a derivation
$(s,\mathit{id}_p(\hat{\gamma}(e_1),\dots,\hat{\gamma}(e_n)))\trans^*(s,\mathit{id}_p(v_1,\dots,\hat{\gamma}(e_n)))\trans^*\dots\trans^*(s,\mathit{id}_p(v_1,\dots,v_n))\trans(s,\delta_p(s,v_1,\dots,v_n))$
by repeated applications of (Red PCon 1) and (Red PCon 2), and a final
application of (Red Pcon 3). (Alternatively, a derivation that reduces
to $\star$ is also possible.) Therefore,
$\mathit{id}_p(\hat{\gamma}(e_1),\ldots,\hat{\gamma}(e_n))\in
R_{\typ{Bool}}$, as required.

\citem{$e=\mathit{id}_a(e_1,\ldots,e_n)$} This case is exactly like
the case for $\mathit{id}_p(e_1,\dots,e_n)$, replacing $\typ{Bool}$ by 
$\typ{Act}$ where appropriate. 

\citem{$e=e_1~e_2$} Assume $\Gamma\vdash e_1~e_2:\tau$, and
$\gamma\sat\Gamma$. We need to show that
$\hat{\gamma}(e_1~e_2)=\hat{\gamma}(e_1)~\hat{\gamma}(e_2)\in
R_\tau$. Since $\Gamma\vdash e_1~e_2:\tau$, we know that $\Gamma\vdash 
e_1:\tau'\rightarrow\tau$, and $\Gamma\vdash e_2:\tau'$, for some
$\tau'$. Applying the induction hypothesis, we get
$\hat{\gamma}(e_1)\in R_{\tau'\rightarrow\tau}$ and
$\hat{\gamma}(e_2)\in R_{\tau'}$. By the definition of
$R_{\tau'\rightarrow\tau}$, we get that
$\hat{\gamma}(e_1)~\hat{\gamma}(e_2)\in R_{\tau}$, as required. 

\citem{$e=e_1?e_2\mo{:}e_3$} Assume $\Gamma\vdash
e_1?e_2\mo{:}e_3:\tau$, and $\gamma\sat\Gamma$. We need to show that
$\hat{\gamma}(e_1?e_2\mo{:}e_3)=\hat{\gamma}(e_1)?\hat{\gamma}(e_2)\mo{:}\hat{\gamma}(e_3)\in
R_{\tau}$. Since $\Gamma\vdash e_1?e_2\mo{:}e_3:\tau$, we know that
$\Gamma\vdash e_1:\typ{Obj}$, $\Gamma\vdash e_2:\tau$, and
$\Gamma\vdash e_3:\tau$, for some $\tau$. Applying the induction
hypothesis, we get $\hat{\gamma}(e_1)\in R_{\typ{Bool}}$,
$\hat{\gamma}(e_2)\in R_\tau$, and $\hat{\gamma}(e_3)\in
R_\tau$. Therefore, for all $s$, we can construct either the
derivation
$(s,\hat{\gamma}(e_1)?\hat{\gamma}(e_2)\mo{:}\hat{\gamma}(e_3))\trans^*(s,\con{true}?\hat{\gamma}(e_2)\mo{:}\hat{\gamma}(e_3))\trans(s,\hat{\gamma}(e_2))\trans^*(s,v_2)$
or the derivation
$(s,\hat{\gamma}(e_1)?\hat{\gamma}(e_2)\mo{:}\hat{\gamma}(e_3))\trans^*(s,\con{false}?\hat{\gamma}(e_2)\mo{:}\hat{\gamma}(e_3))\trans(s,\hat{\gamma}(e_3))\trans^*(s,v_3)$,
using (Red If 1), (Red If 2), (Red If 3), depending on the Boolean
value that $(s,\hat{\gamma}(e_1))$ reduces to. (Alternatively, a
derivation that reduces to $\star$ is also possible.) Therefore,
$\hat{\gamma}(e_1)?\hat{\gamma}(e_2)\mo{:}\hat{\gamma}(e_3)\in
R_{\tau}$, as required.

\citem{$e=e_1;e_2$} Assume $\Gamma\vdash e_1;e_2:\typ{Act}$, and
$\gamma\sat\Gamma$. We need to show that
$\hat{\gamma}(e_1;e_2)=\hat{\gamma}(e_1);\hat{\gamma}(e_2)\in
R_{\typ{Act}}$. Since $\Gamma\vdash e_1;e_2:\typ{Act}$, we know that
$\Gamma\vdash e_1:\typ{Act}$ and $\Gamma\vdash
e_2:\typ{Act}$. Applying the induction hypothesis, we get
$\hat{\gamma}(e_1)\in R_{\typ{Act}}$ and $\hat{\gamma}(e_2)\in
R_{\typ{Act}}$. Therefore, for all $s$, we can construct the
derivation
$(s,\hat{\gamma}(e_1);\hat{\gamma}(e_2))\trans^*(s',\con{skip};\hat{\gamma}(e_2))\trans(s',\hat{\gamma}(e_2))\trans^*(s'',\con{skip})$,
by applications of (Red Seq 1), (Red Seq 2), (Red Seq 3).
(Alternatively, a derivation that reduces to $\star$ is also
possible.) Therefore, we have $\hat{\gamma}(e_1);\hat{\gamma}(e_2)\in
R_{\typ{Act}}$, as required. 
\end{itemize}

If $\Jpure{\tau}$, a straightforward induction on the length of the
derivation $(s,e)\trans^*(s',v)$, via Lemma~\ref{l:third}, establishes
that $s'=s$. 
\end{proof}

\begin{namedtheorem}{Theorem}{\ref{t:safe-trans}}
If $\alpha_1:A_1,\dots,\alpha_n:A_n\Rightarrow \alpha:A$ has a
derivation that respects imperative structure, then the rule
\[ \Rule{\Gamma\vdash\alpha_1:T(A_1)\quad\dots\quad
\Gamma\vdash\alpha_n:T(A_n)}{\Gamma\vdash\alpha:T(A)} \]
is an admissible typing rule.   
\end{namedtheorem}

\begin{proof}
We proceed by induction on the height of the derivation for
$\alpha_1:A_1,\dots,\alpha_n:A_n\Rightarrow \alpha:A$. First, some
notation: if $\Delta$ is a sequence $\alpha_1:A_1,\dots,\alpha_k:A_k$
and $\Gamma$ is a typing context, we write $\Gamma\intension{\Delta}$
for the sequence of judgments
$\Gamma\vdash\alpha_1:T(A_1),\dots,\Gamma\vdash\alpha_k:T(A_k)$. For
the base case, we have $\alpha:A\Rightarrow \alpha:A$, and clearly,
the typing rule \[\Rule{\Gamma\vdash
\alpha:T(A)}{\Gamma\vdash\alpha:T(A)}\] is admissible. For the
induction step, consider a number of cases, one for each possible last
rule of the derivation. In the case (Seq Cut), the last rule of the
derivation is of the form
\[
\Rule{\Delta\Rightarrow\beta: B & 
      \Sigma_1,\beta:B,\Sigma_2\Rightarrow\alpha:A}
  {\Sigma_1,\Delta,\Sigma_2\Rightarrow\alpha:A.}
\]
Applying the induction hypothesis, both
\[
\Rule{\Gamma\intension{\Delta}}
     {\Gamma\vdash\beta:T(B)}\]
and
\[
\Rule{\Gamma\intension{\Sigma_1} & \Gamma\vdash\beta:T(B) & \Gamma\intension{\Sigma_2}}
     {\Gamma\vdash\alpha:T(A)}
\]
are admissible typing rules. Composing these two admissible rules
yields the admissible rule:
\[
  \Rule{\Gamma\intension{\Sigma_1}& 
        \Rule{\Gamma\intension{\Delta}}
             {\Gamma\vdash\beta:T(B)}
        & \Gamma\intension{\Sigma_2}}
       {\Gamma\vdash\alpha:T(A).}
\]

In the case (Seq App Right), the last rule of the derivation is of the form
\[
\Rule{\Delta\Rightarrow\beta:B & 
      \Sigma_1,\gamma(\beta):C,\Sigma_2\Rightarrow\alpha:A}
     {\Sigma_1,\gamma:C/B,\Delta,\Sigma_2\Rightarrow\alpha:A}
\]
Applying the induction hypothesis, both
\[ 
\Rule{\Gamma\intension{\Delta}}
     {\Gamma\vdash\beta:T(B)}
\]
and
\[ 
\Rule{\Gamma\intension{\Sigma_1} & 
      \Gamma\vdash\gamma(\beta):C & 
      \Gamma\intension{\Sigma_2}}
     {\Gamma\vdash\alpha:T(A)}
\]
are admissible typing rules. Composing them yields the following
admissible rule, upon noting that $T(C/B)=T(B)\rightarrow T(C)$:
\[
\Rule{\Gamma\intension{\Sigma_1} & 
      \Rule{\Gamma\vdash\gamma:T(B)\rightarrow T(C) &
            \Rule{\Gamma\intension{\Delta}}
                 {\Gamma\vdash\beta:T(B)}}
           {\Gamma\vdash\gamma(\beta):T(C)} &
      \Gamma\intension{\Sigma_2}}
     {\Gamma\vdash\alpha:T(A).}
\]
The case for (Seq App Left) is similar.

Finally, in the case (Seq Abs Right), where we have $\alpha=\lambda
x.\beta$ and $A=B/C$, the last rule of the derivation is of the form
\[
\Rule{\Sigma,x:C\Rightarrow \beta:B}
     {\Sigma\Rightarrow\lambda x.\beta:B/C}
\]
Applying the induction hypothesis, for $\Gamma$ of the form
$\Gamma',x:T(C)$, the typing rule
\[
\Rule{(\Gamma',x:T(C))\intension{\Sigma} &
      \Gamma',x:T(C)\vdash x:T(C)}
     {\Gamma',x:T(A)\vdash\beta:T(B)}
\]
is admissible. Noting that $T(B/C)=T(C)\rightarrow T(B)$, in order to
derive an admissible typing rule using (Typ Fun), we need to check
that $\Jok{T(C)\rightarrow T(B)}$. 
But this is exactly what the assumption that the derivation respects
imperative structure gives us. 
We can therefore derive the following admissible typing rule:
\[ 
\Rule{\Rule{(\Gamma',x:T(C))\intension{\Sigma} &
            \Rule{}
                 {\Gamma',x:T(C)\vdash x:T(C)}}
           {\Gamma',x:T(C)\vdash\beta:T(B)} & \Jok{T(C)\rightarrow T(B)}}
     {\Gamma'\vdash\lambda x\mo{:}T(C).\beta:T(C)\rightarrow T(B).}
\]
The case for (Seq Abs Left) is similar. 
\end{proof}

\bibliographystyle{chicagor}
\bibliography{riccardo2}

\end{document}